\documentclass[journal]{IEEEtran}

\ifCLASSINFOpdf
\else
   \usepackage[dvips]{graphicx}
\fi
\usepackage{url}

\usepackage{amsthm}
\usepackage{amsmath}
\usepackage[ruled,vlined]{algorithm2e}
\usepackage{amssymb}
\usepackage{subcaption}
\usepackage{tabularx}
\usepackage{graphicx}
\usepackage{bm,bbm}
\usepackage{color}
\usepackage{tabularx}
\usepackage{multirow}
\usepackage{diagbox}
\usepackage{nicefrac}

\newcommand{\y}{\mathbf{y}}
\newcommand{\bias}{\mathbf{a}}
\newcommand{\z}{\mathbf{z}}
\newcommand{\x}{\mathbf{x}}
\newcommand{\h}{\mathbf{h}}
\newcommand{\g}{\mathbf{g}}
\newcommand{\T}{\text{T}}
\newcommand{\bigs}{\mathbf{S}}

\newcommand{\bige}{\mathbf{E}}
\newcommand{\F}{\mathbf{F}}
\newcommand{\X}{\mathbf{X}}
\newcommand{\W}{\mathbf{W}}
\newcommand{\I}{\mathbf{I}}
\newcommand{\loss}{\mathcal{L}(\mathbf{y})}
\newcommand{\boldmu}{\pmb{\mu}}
\newcommand{\boldeta}{\pmb{\eta}}

\begin{document}

\title{Gaussian Process Convolutional Dictionary Learning}

\author{Andrew H. Song, \IEEEmembership{Student member, IEEE}, Bahareh Tolooshams, \IEEEmembership{Student member, IEEE},\\ and Demba Ba, \IEEEmembership{Member, IEEE}
\thanks{Andrew H. Song is with the Electrical Engineering and Computer Science,
Massachusetts Institute of Technology, Cambridge, MA 02139 USA (e-mail:
andrew90@mit.edu).}
\thanks{Bahareh Tolooshams and Demba Ba are with the School of Engineering and Applied Sciences, Harvard
University, Allston, MA 02134 USA (e-mail:
btolooshams@seas.harvard.edu; demba@seas.harvard.edu).}}

\maketitle

\begin{abstract}
Convolutional dictionary learning (CDL), the problem of estimating shift-invariant templates from data, is typically conducted in the absence of a prior/structure on the templates. In data-scarce or low signal-to-noise ratio (SNR) regimes, learned templates overfit the data and lack smoothness, which can affect the predictive performance of downstream tasks. To address this limitation, we propose GPCDL, a convolutional dictionary learning framework that enforces priors on templates using Gaussian Processes (GPs). With the focus on smoothness, we show theoretically that imposing a GP prior is equivalent to Wiener filtering the learned templates, thereby suppressing high-frequency components and promoting smoothness. We show that the algorithm is a simple extension of the classical iteratively reweighted least squares algorithm, independent of the choice of GP kernels. This property allows one to experiment flexibly with different smoothness assumptions. Through simulation, we show that GPCDL learns smooth dictionaries with better accuracy than the unregularized alternative across a range of SNRs. Through an application to neural spiking data, we show that GPCDL learns a more accurate and visually-interpretable smooth dictionary, leading to superior predictive performance compared to non-regularized CDL, as well as parametric alternatives.
\end{abstract}

\textit{This manuscript is an extended version of the IEEE Signal Processing
	Letters paper (doi:10.1109/LSP.2021.3127471), with the supplementary material as the appendix.}\\


\IEEEpeerreviewmaketitle

\section{Introduction}
In recent years, the practice of modeling signals as a combination of a few repeated templates has gained popularity~\cite{Lewicki99}, such as in the modeling of point spread functions for molecular~\cite{Betzig06} and astronomical imaging~\cite{wohlberg21}, or action potentials in biological signals~\cite{Lewicki98, Vogelstein10, Song20}. This is referred to as convolutional dictionary learning (CDL), where the goal is to estimate the shape, locations, and amplitudes of the shift-invariant templates~\cite{Cardona18}. The dictionary (the collection of the templates) is usually learned in a data-driven manner, without constraints.

In practice, when data are scarce or have a low signal-to-noise ratio (SNR), learned dictionaries overfit the data in the absence of constraints. Consequently, the interpretability of the dictionary and its predictive performance on unobserved data suffer. The problem is aggravated for data from non-Gaussian distributions such as binomial data, due to the non-linear mapping from dictionary to observations~\cite{Giryes14}. There is also evidence that the templates for naturally-occurring data could be considered \textit{smooth}~\cite{wohlberg21, Lewicki98}. 

The recent literature suggests that there are several approaches to learning smooth shift-invariant templates. One approach models the templates with parametric functions, such as the bi-exponential~\cite{Vogelstein10} function or a mixture of Gaussians~\cite{Sadras19}. Another line of work imposes total variation or Tikhonov-like penalties~\cite{Huo13, dohmatob16,Yan12} on the templates. More recently, smooth templates were obtained by passing learned dictionary through pre-designed lowpass filters~\cite{wohlberg21, Song20, Soh21}.

We propose an alternative flexible, nonparametric approach, by assuming that the templates are generated from a Gaussian Process (GP)~\cite{GP}. We make the following contributions\footnote{The code can be found at https://github.com/andrewsong90/gpcdl} \\
\textbf{CDL via GP regularization} We introduce GPCDL, a framework for CDL with GP regularization, which can be applied to observations from the natural exponential family~\cite{glm}. We show that the learned dictionary is accurate in conditions where the unregularized alternatives overfit.
The learning procedure is a simple extension of iteratively reweighted least squares and allows us to easily incorporate the GP prior.\\	
\textbf{GP prior as Wiener filter} We show that, under some assumptions, the GP prior acts as a lowpass Wiener filter~\cite{Wiener64}, which allows GPCDL to learn smooth dictionaries. From this unique perspective, we elucidate the trade-off between the amount of training data and the parameters of the GP prior.

The paper is organized as follows: Section~\ref{section:background} and \ref{section:inference} introduce the background and the GPCDL framework. Section~\ref{section:analysis} develops the interpretation of GPCDL as Wiener filtering. In Section~\ref{section:experiments} and \ref{section:conclusion}, the results and conclusion are presented.

\section{Background}\label{section:background}
\subsection{Notation}
We denote the zero and identity matrices as $\mathbf{0}$ and $\mathbf{I}$, with appropriate dimensions. $\mathbf{A}_{(k,k')}$ refers to the entry of matrix $\mathbf{A}$ at location $(k,k')$. The $\operatorname{diag}(\cdot)$ refers to a diagonal matrix, with entries equal to the vector argument. When applied to a vector, a function operates in an element-wise manner. 

\subsection{Natural exponential family}
Let $\y^j\!\in\!\mathbb{R}^N$ be observations from the natural exponential family with mean $\boldmu^j=\mathbb{E}[\y^j]$, for $j=1,\ldots,J$. With $\mathbf{1}_N$ as the $N$-length vector of ones, the log-likelihood is given as
\begin{equation}\label{eq:likelihood}
\log \ell(\y^j)= \frac{f(\boldmu^j)^{\T}\y^j - \mathbf{1}_N^{\text{T}}b(f(\boldmu^j))}{\phi} + c(\y^j,\phi),
\end{equation}
where $\phi$ is a dispersion parameter and the functions $b(\cdot)$, $c(\cdot)$, as well as the invertible link $f(\cdot)$, are distribution-dependent. 

We consider $f(\boldmu^j)$ to be the sum of scaled and time-shifted copies of $C$ finite-length templates $\{\mathbf{h}_c\}_{c=1}^C\in\mathbb{R}^K$, each localized, i.e., $K$ $\ll$ $N$. We express $f(\boldmu^j)$ as a convolution, i.e., $f(\boldmu^j)=\sum_{c=1}^C \h_c\ast\x_c^j+\bias^j$, where the \textit{code vector} $\x_c^j\in\mathbb{R}^{N-K+1}$ is a train of scaled impulses and $\bias^j\in\mathbb{R}^N$ is a baseline. The entry of $\x_c^j$ at index $n_{c,i}^j$ corresponds to the location of the $i^{th}$ event with amplitude $x_{c,i}^j$. Alternatively, we can write $f(\boldmu^j)-\bias^j=\sum_c\X^j_c\h_c=\sum_c\sum_{i=1}^{N_c^j}x^j_{c,i}\bigs^j_{c,i}\h_c$, where $\bigs^j_{c,i}=[\mathbf{0}_{K\times (n^{j}_{c,i}-1)}\,\, \mathbf{I}_{K\times K}\,\, \mathbf{0}_{K\times (N-K-n^j_{c,i}+1)}]^{\T}\in\mathbb{R}^{N\times K}$ is the linear operator that shifts $\h_c$ by $n^j_{c,i}$ samples and $N_c^j$ is the number of occurrences of $\h_c$ in $\y^j$~\cite{Song20}.

\subsection{Gaussian Process}
Gaussian Processes (GPs) offer a nonparameteric and flexible Bayesian approach for signal modeling~\cite{GP}, which we use as a smooth prior on $\h_c$. We first define functions $h_c:[0, T) \to \mathbb{R}$, $\forall c$, generated from a GP prior with zero-mean and stationary kernel $\kappa_c(t,t')$, i.e., $h_c(t)\sim \operatorname{GP}(0, \kappa_c(t,t')),\forall c$. We assume that the filter $\h_c$ is sampled from $h_c(\cdot)$, and for simplicity, with constant sampling interval $\Delta$ such that $T=K\Delta$. This yields $\h_c\sim \mathcal{N}(\mathbf{0}, \Sigma_c)$, where $\Sigma_c\in\mathbb{R}^{K\times K}$ is the covariance matrix and $\Sigma_{c, (k,k')}=\kappa_c(k\Delta, k'\Delta)$.

We focus on kernels in the \textit{Matern} family~\cite{Matern60}, parameterized by $\nu$, variance $\sigma_c^2$, and lengthscale $\l_c$. The parameter $\nu$ controls the smoothess of the kernel and is defined \textit{a priori} by the user. The popular choice is $\nu=p+1/2$, $p\in\mathbb{N}^{+}$, since this leads to simplification of the kernel expression. The parameters $\sigma_c^2$ and $l_c$ can be chosen by maximum-likelihood estimation or cross-validation~\cite{GP}. 

The power spectral density (PSD) of the kernel, denoted $\gamma_c(\omega)$, a function of the normalized frequency $\omega\in[-\pi,\pi]$, is obtained by taking the Fourier transform of the kernel~\cite{Bochner59}. We focus on $\nu=1.5$ throughout this work, noting that the same holds for any other GP kernels. For $\nu=1.5$, we have 
\begin{equation*}
\begin{split}
\Sigma_{c, (k,k')} &= \sigma_c^2\Big(1+\dfrac{\sqrt{3}(k-k')\Delta}{l_c}\Big)\exp\Big(-\sqrt{3}\dfrac{(k-k')\Delta}{l_c}\Big)\\
\gamma_c(\omega)&=(4/\sqrt{3})\sigma_c^2l_c/(1+l_c^2\omega^2/3)^{2}.\\
\end{split}
\end{equation*}
An example of $\gamma_c(\omega)$ for $\nu=1.5$ is depicted in Fig.~\ref{fig:gain}(a) for varying $l_c$. As $\omega$ increases, $\gamma_c(\omega)$ decays monotonically.

\section{CDL with GP regularization}\label{section:inference}
\subsection{Objective}
Combining the log-likelihood $\log p(\y|\{\h_c\})$, where we use $\y$ to denote $\{\y^j\}$, and the log-prior, we cast the GPCDL problem as minimizing the negative \textit{log-posterior} $\loss$,
\begin{equation}\label{eq:secdl}
\begin{split}
\min_{\substack{\{\mathbf{h}_c\}_{c=1}^C\\ \{\x_c^j\}_{c=1,j=1}^{C,J}}}& \overbrace{\underbrace{\sum_j\frac{-f(\boldmu^j)^{\T} \y^j + \mathbf{1}_N^{\text{T}}b(f(\boldmu^j))}{\phi}}_{-\log p(\y|\{\h_c\})}+\sum_c\frac{\h_c^{\T}\Sigma_c^{-1}\h_c}{2}}^{\loss}\\
&\text{s.t.}\quad \lVert \x_c^j\rVert_0<\beta \text{ and } \lVert\h_c\rVert_2\leq 1, \forall j,c.\\
\end{split}
\end{equation}

We use the $\ell_0$ pseudo-norm for the sparsity constraint (number of nonzeros), with sparsity level $\beta$. The GP prior is incorporated as a quadratic regularizer on $\h_c$.  This formulation can be naturally extended to the multivariate setting.

Parametric approaches express $\{\h_c\}$ as combinations of parametric functions~\cite{Vogelstein10,Sadras19}. Despite requiring few parameters, these approaches require a careful choice of functions and parameters (e.g., the number of functions) to minimize model misspecification error. GPCDL is a nonparametric approach and avoids the misspecification issue at the expense of more parameters, i.e., the templates. By imposing structure on $\{\h_c \}$ with the GP prior, GPCDL promotes smooth templates, while maintaining the flexibility of the nonparametric paradigm.

We use alternating minimization to solve Eq.~(\ref{eq:secdl}), where $\loss$ is minimized with respect to $\{\h_c\}$ and $\{\x_c^j\}$, by alternating between a convolutional sparse coding (CSC) step, optimizing for $\{\x_c^j\}$, and a convolutional dictionary update (CDU) step, optimizing for $\{\h_c\}$~\cite{Cardona18}. For CSC, we use Convolutional Orthogonal Matching Pursuit (COMP)~\cite{Lozano2011, Tolooshams20}, a greedy algorithm that iteratively identifies 
the template and the code that minimize $-\log p(\y|\{\h_c\})$. We define $\beta$ as the \textit{minimal} active number of elements which, when reconstructed in the form of $\boldmu$, results in $-\log p(\y|\{\h_c\})$ lower than a threshold computed from the baseline period of each dataset. More details can be found in~\cite{Tolooshams20}. 

\subsection{Convolutional Dictionary Update} Given the estimates for $\X_c^j$, we use Newton's method to minimize $\loss$ with respect to $\h_c$, referred to, in the context of the exponential family, as iteratively reweighted least squares (IRLS)~\cite{fahrmeir2013multivariate}. At iteration $t$, we compute its gradient and Hessian
\begin{align}
\nabla_{\h_c}\loss&=-\phi^{-1}\sum_j(\X_c^j)^{\T}(\y^j-\boldmu^{j,(t)}) + \Sigma_c^{-1}\h_c^{(t)},\\
\nabla^2_{\h_c}\loss&=\phi^{-1}\sum_j(\X_c^j)^{\T}\operatorname{diag}((f'(\boldmu^{j,(t)}))^{-1})\X_c^j + \Sigma_c^{-1},\notag
\end{align}
where $f'$ denotes the derivative of $f$. Letting $\W_c^{j,(t)}=\operatorname{diag}((f'(\boldmu^{j,(t)}))^{-1})$, we have
\begin{align}\label{eq:irls}
&\h_c^{(t+1)} =\h_c^{(t)}-(\nabla_{\h_c}^2\loss)^{-1}\nabla_{\h_c}\loss\\
&=(\phi^{-1}\sum_{j'}(\X^{j'}_c)^{\T}\W_c^{j',(t)}\X_c^{j'}+\Sigma_c^{-1})^{-1}\sum_j(\X_c^j)^{\T}\z_c^{j,(t+1)},\notag
\end{align}
where $\z_c^{j,(t+1)} = \phi^{-1}(\W_c^{j,(t)}\X_c^j\h_c^{(t)} + (\y^j - \boldmu^{j,(t)}))\in\mathbb{R}^N$. After each update, we normalize $\h_c^{(t+1)}$ to have unit norm. We update $\h_c$ in a cyclic manner and proceed to the next CSC step. The role of $(\X_c^j)^{\T}$ in $(\X_c^j)^{\T}\z_c^{j,(t+1)}$ is to extract the segments of $\z_c^{j, (t+1)}$ where $\h_c^{(t)}$ occurs, and take their weighted average~\cite{Song20}. Since $K\ll N$, the computational complexity of matrix inversion for $\Sigma_c$ and $\nabla_{\h_c}^2\loss$ is negligible.

In summary, the CDU step seamlessly incorporates the GP constraint into the classical IRLS algorithm~\cite{glm}. Since the optimization is not dependent on the form of $\Sigma_c$, we can choose different $\Sigma_c$ to enforce different degrees of smoothness. This is simpler compared to approaches utilizing total-variation like penalties~\cite{Huo13, dohmatob16}, which require custom, dedicated primal-dual optimization methods for different penalties~\cite{Chambolle2009AnIT}. 

\section{Analysis of converged dictionary}\label{section:analysis}
We now analyze how GPCDL promotes the smoothness of $\h_c$. We focus mainly on the case where the observations are Gaussian for intuition. We assume that the templates are non-overlapping, that is $(\bigs^j_{c,i})^{\T}\bigs^j_{c,i'}=\mathbf{0}$ for $i\neq i'$.

\noindent \textbf{Gaussian case} IRLS converges in a single iteration (we omit the index $t$), with $f$ as the identity and $\W_c^{j}=\mathbf{I}_{N\times N}$. This yields $\z_c^j=\phi^{-1}(\y^j-\sum_{c'\neq c}\X_{c'}^j\h_{c'}-\bias^j)$. The dispersion is the observation noise variance, i.e., $\phi=\sigma_{\varepsilon}^2$.
\begin{equation}
\begin{split}
\h_c&=\Big(\sigma_{\varepsilon}^{-2}\sum_j(\X_c^{j})^{\T}\X_c^j + \Sigma_c^{-1}\Big)^{-1}\sum_{j}(\X_c^{j})^{\T}\z_c^j\\
&=\Big(\sigma_{\varepsilon}^{-2}\sum_{j,i}(x_{c,i}^j)^2\mathbf{I}+ \Sigma_c^{-1}\Big)^{-1}\sum_{j}(\X_c^{j})^{\T}\z_c^j,\\
\end{split}
\end{equation}
where the second equality follows from $(\bigs^j_{c,i})^{\T}\bigs^j_{c,i'}=\mathbf{0}$ for $i=i'$. The factor $\alpha^2=\sigma_{\varepsilon}^{-2}\sum_{j,i}(x_{c,i}^j)^2$, which we term \textit{code-SNR}, represents the SNR of the sparse codes, since $\sum_{j,i}(x_{c,i}^j)^2$ and $\sigma_{\varepsilon}^{2}$ are the energy of the codes and the noise, respectively. 

Let us now examine $\F\h_c$, the spectra of $\h_c$,  where $\F\in\mathbb{C}^{K\times K}$ is a discrete Fourier transform matrix, with $\F_{k,k'}=\exp(-2\pi i(k-1)(k'-1)/K)$, and $\omega_k=2\pi k/K$. Using the eigen-decomposition for a stationary kernel~\cite{Turner2014}, we get
\begin{equation}
\Sigma_c \simeq \F^{-1}\operatorname{diag}([\gamma_c(\omega_1), \ldots, \gamma_c(\omega_K)])\F.
\end{equation}
Denoting $\bige_c=\sum_{j}(\X_c^{j})^{\T}\z_c^j$ for notational simplicity, and using $\F\F^{-1}=\I$, we have
\begin{align}\label{eq:wf}
\F\h_c&\simeq\F(\alpha^2\mathbf{I}+\F^{-1}\operatorname{diag}([\gamma_c^{-1}(\omega_1), \ldots, \gamma^{-1}_c(\omega_K)])\F)^{-1}\bige_c\notag\\
&=\operatorname{diag}\big(\g\big)\F\widetilde{\h}_c,
\end{align}
where $\g_k=\gamma_c(\omega_k)/(\gamma_c(\omega_k)+\alpha^{-2})$ and $\widetilde{\h}_c=\bige_c/\alpha^2.$ We can interpret Eq.~(\ref{eq:wf}) as \textit{Wiener filter}~\cite{Wiener64} with gain $\g_k$ at $\omega=\omega_k$ on $\widetilde{\h}_c$, the learned template \textit{without} the regularization.

\begin{figure}[!t]
	\centering
	\includegraphics[width=\linewidth]{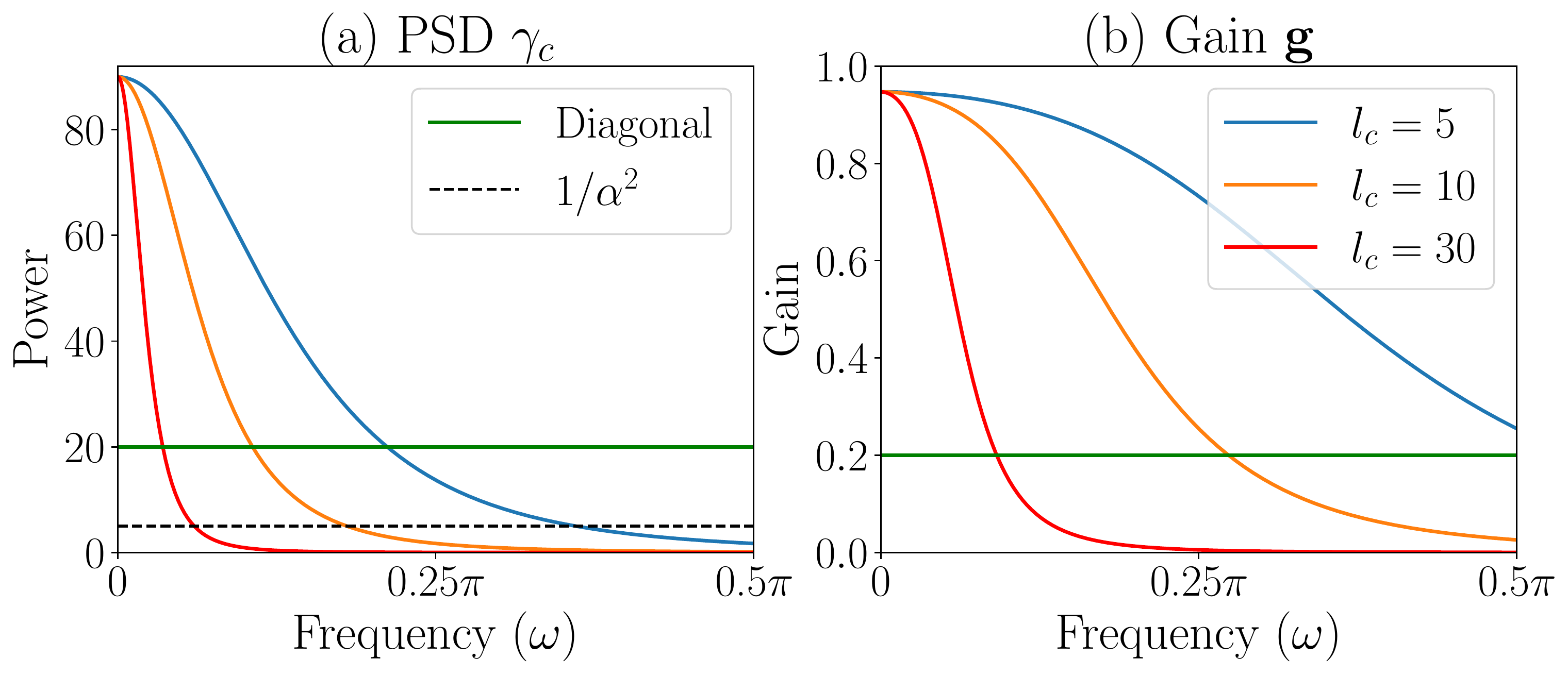}
	\caption{(a) PSD $\gamma_c(\omega)$ for Matern kernel with $\nu=1.5$ and varying $l_c$ for fixed $\alpha$. The green line corresponds to Tikhonov regularization with diagonal $\sigma_c^2$. (b) The filter gain $\mathbf{g}$.}
	\label{fig:gain}
\end{figure}

The gain $\g_k$ depends on two factors: 1) the code-SNR $\alpha^2$ and 2) the PSD of the GP prior $\gamma_c(\omega_k)$. For fixed $\alpha^2$,  the larger (and smaller) $\gamma_c(\omega_k)$, the closer $\g_k$ to 1 (and 0). Therefore, $\g_k$ acts as a lowpass filter and suppresses high-frequency content, allowing accurate learning of smooth $\h_c$. Fig.~\ref{fig:gain} demonstrates how different $l_c$ lead to different gains $\g$. If $\alpha^2$ is increased by collecting more data (increasing $J$), $\g_k$ increases across the entire $\omega$ axis and the filtering effect diminishes. This agrees with the Bayesian intuition that with more data, the likelihood dominates the prior. Note that with increasing $J$, $\widetilde{\h}_c$ itself becomes more accurate~\cite{Agarwal2016LearningSU}.

This suggests that GPCDL shares the same philosophy as~\cite{wohlberg21, Song20}, since the learned dictionary is lowpass-filtered. However, the filters are designed differently. For GPCDL, the Wiener filter is \textit{data-adaptive}, as the gain is determined \textit{a posteriori} from the balance between the likelihood (data) and the prior. In contrast, the filter is designed \textit{a priori} in~\cite{wohlberg21, Song20, Soh21}, without reference to the data or optimization criteria.

We note that a similar form has been studied in the spectral filtering theory for Tikhonov regularization~\cite{Leary01}. Tikhonov regularization can be recovered from Eq.~(\ref{eq:secdl}) with $\Sigma_c=\sigma_c^2\I$. The diagonal covariance yields $\gamma_c(\omega_k)=\gamma_c$, $\forall k$, and consequently constant gain $\g_k=\g$, $\forall k$, resulting in $\widetilde{\h}$ with a smaller norm, shown in Fig.~\ref{fig:gain} (green). For GPCDL, however, $\Sigma_c$ is symmetric and non-diagonal. This allows GPCDL to have frequency-dependent Wiener filter gain.

\noindent \textbf{General case }  For non-Gaussian distributions, two factors complicate the interpretation: 1) IRLS requires multiple iterations to converge and 2) $\W_c^{j,(t)}$ is dependent on $\omega_k$ and iteration $t$. However, we conjecture that smoothing still takes place. Specifically, $\mathbf{R}^{(t)}=(\X_c^j)^{\T}\W_c^{j,(t)}\X_c^j$ is still a diagonal matrix with $\mathbf{R}^{(t)}_{(k,k)}=\phi^{-1}\sum_{j,i}(x^j_{c,i})^2\cdot(f'(\boldmu_{n^j_{c,i}+k-1}^{j,(t)}))^{-1}$. This consequently yields $\g_k = \gamma_c(\omega_k)/(\gamma_c(\omega_k)+(\mathbf{R}^{(t)}_{(k,k)})^{-1})$, computed using $\mathbf{R}^{(t)}=\F^{-1}\mathbf{R}^{(t)}\F$. Therefore, the relation between $\gamma_c(\omega_k)$ and $(\mathbf{R}^{(t)}_{(k,k)})^{-1}$ holds as in the Gaussian case. Consequently, $\g_k$ filters the spectra of weighted-averaged segments from $\z_c^{j,(t)}$, extracted by the operator $(\X_c^j)^{\T}$. Empirically, we observe that low-pass filtering still occurs.

\section{Experiments}\label{section:experiments}
We apply our framework to two datasets: 1) simulated data (Gaussian) and 2) neural spiking data from rats (Bernoulli). We use the Matern kernel with $\nu=1.5$, fix $\sigma_c^2=1$, and vary $l_c$ to control the regularization. We use the mixture of Gaussians (MOG) model $\h^{\text{MOG}}_c[k]=\sum_{d=1}^D a_{c,d}\exp(-(k-\mu_{c,d})^2/\sigma_{c,d}^2)$ as baseline, with parameters $\{a_d, \mu_d, \sigma_d^2 \}_{d=1}^D$ determined by maximum-likelihood estimation. MOG represents a smooth parametric approach. We run 15 iterations of our algorithm, with $\widehat{\h}_c$ and $\widehat{\x}_c$ denoting the solutions at convergence.

\subsection{Simulated data}
\begin{figure}[!t]
	\centering
	\includegraphics[width=0.95\linewidth]{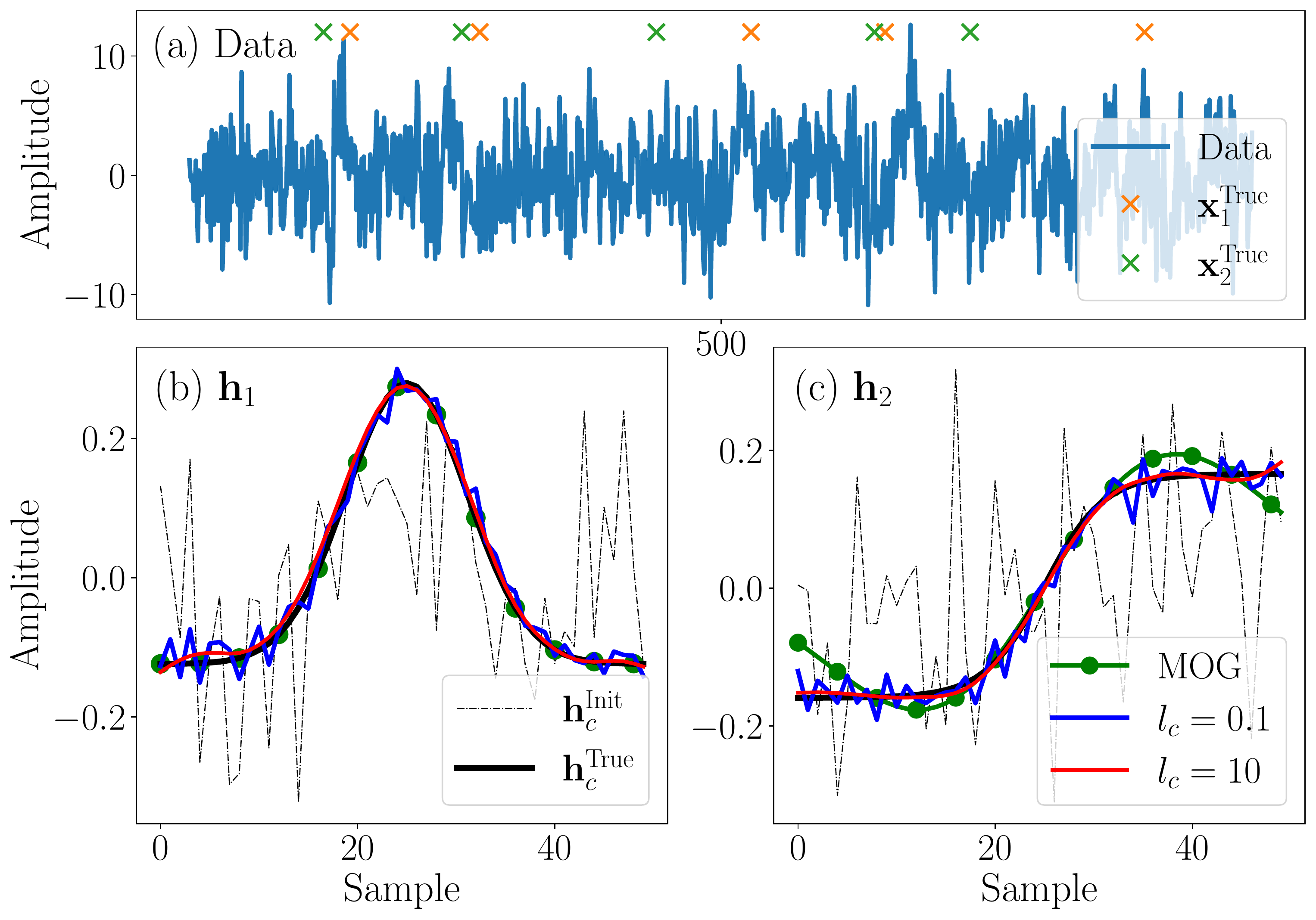}
	\caption{Simulation results for $J\!=\!100$ and $\sigma_{\varepsilon}^2=10$. (a) An example data trace and true codes. (b-c) Dictionary elements.}
	\label{fig:simulation}
\end{figure}

\noindent \textbf{Dataset} We simulated Gaussian data with $\{\h_c^{\text{True}} \}_{c=1}^2\!\in\!\mathbb{R}^{50}$ (Fig.~\ref{fig:simulation} (black) Gaussian and sigmoid), each appearing 4 times with magnitude uniformly sampled from $[10,20]$, throughout the length $N\!=\!1{,}000$ signal. 
The signal is perturbed with Gaussian noise  with variance $\sigma_{\varepsilon}^2=5$. For evaluation, we use the dictionary error, $\operatorname{err}(\widehat{\h}_c)=(1-\langle\widehat{\h}_c, \h_c^{\text{True}}\rangle^2)^{\frac{1}{2}}$~\cite{Agarwal2016LearningSU}. We perturbed $\h_c^{\text{True}}$ with Gaussian noise and obtain $\h_c^{\text{Init}}$ (dotted black) with $\operatorname{err}(\h_c^{\text{Init}})>0.7$. We averaged the power $\omega\in[0.5\pi, \pi]$ to obtain the dispersion $\widehat{\phi}=\widehat{\sigma}_{\varepsilon}^2$.

\noindent \textbf{Results} Table~\ref{table:simulation} shows the error, averaged over 10 independent runs, for varying SNR and lengthscale $l_c$. The larger the $l_c$, the stronger the GP regularization, resulting in considerably lower errors, as visually supported in Fig.~\ref{fig:simulation}. The learned $\widehat{\h}_c$ for $l_c\!=\!0.1$ (blue) corresponding to minimal regularization, contains high-frequency noise. With GP regularization ($l_c\!=\!10$, red), the noise is filtered out, and thus $\widehat{\h}_c$ is more accurate with the same code-SNR. As expected, the overall errors are lower with higher code-SNR, where $(J, \sigma_{\varepsilon}^2)\!=\!(10,10)$ and $(100,5)$ correspond to the lowest and the highest code-SNR. Even with high code-SNR, we observe the benefits of GP regularization.

Figs.~\ref{fig:simulation} (b-c) also depict $\widehat{\h}^{\text{MOG}}_1$ and $\widehat{\h}^{\text{MOG}}_2$, optimized with $D=1$ and 2, respectively. This shows potential issues of model misspecification in the parametric approach, as observed in Fig.~\ref{fig:simulation} (c), where $\widehat{\h}^{\text{MOG}}_2$ cannot adequately model the sigmoid. On the other hand, the nonparametric GPCDL does not face this issue. 

\begin{table}[!t]
	\caption{Dictionary error $\operatorname{err}(\widehat{\h}_c)$ for simulated data with $J=\{10,100\}$ and $\sigma_{\varepsilon}^2=\{5,10\}$.}
	\label{table:simulation}
	\centering
	\begin{tabular}{|c|c|c|c|c|c|c|c|}
		\hline
		&$l_c$&\multicolumn{2}{c}{$0.1$}&\multicolumn{2}{|c}{$25$}&\multicolumn{2}{|c|}{$100$}\\
		\hline
		Error&\diagbox{$\sigma_{\varepsilon}^2$}{$J$}&10&100&10&100&10&100\\
		\hline
		$\operatorname{err}(\widehat{\h}_1)$&5&0.29&0.18&0.18&0.12&\textbf{0.13}&\textbf{0.06}\\
		$\operatorname{err}(\widehat{\h}_1)$&10&0.45&0.30&0.36&0.23&\textbf{0.20}&\textbf{0.11}\\
		\hline
		$\operatorname{err}(\widehat{\h}_2)$&5&0.32&0.18&0.21&0.11&\textbf{0.10}&\textbf{0.06}\\
		$\operatorname{err}(\widehat{\h}_2)$&10&0.46&0.31&0.28&0.24&\textbf{0.17}&\textbf{0.14}\\
		\hline
	\end{tabular}
\end{table}

\subsection{Neural activity data from barrel cortex}
\begin{figure}[!t]
	\centering
	\includegraphics[width=0.95\linewidth]{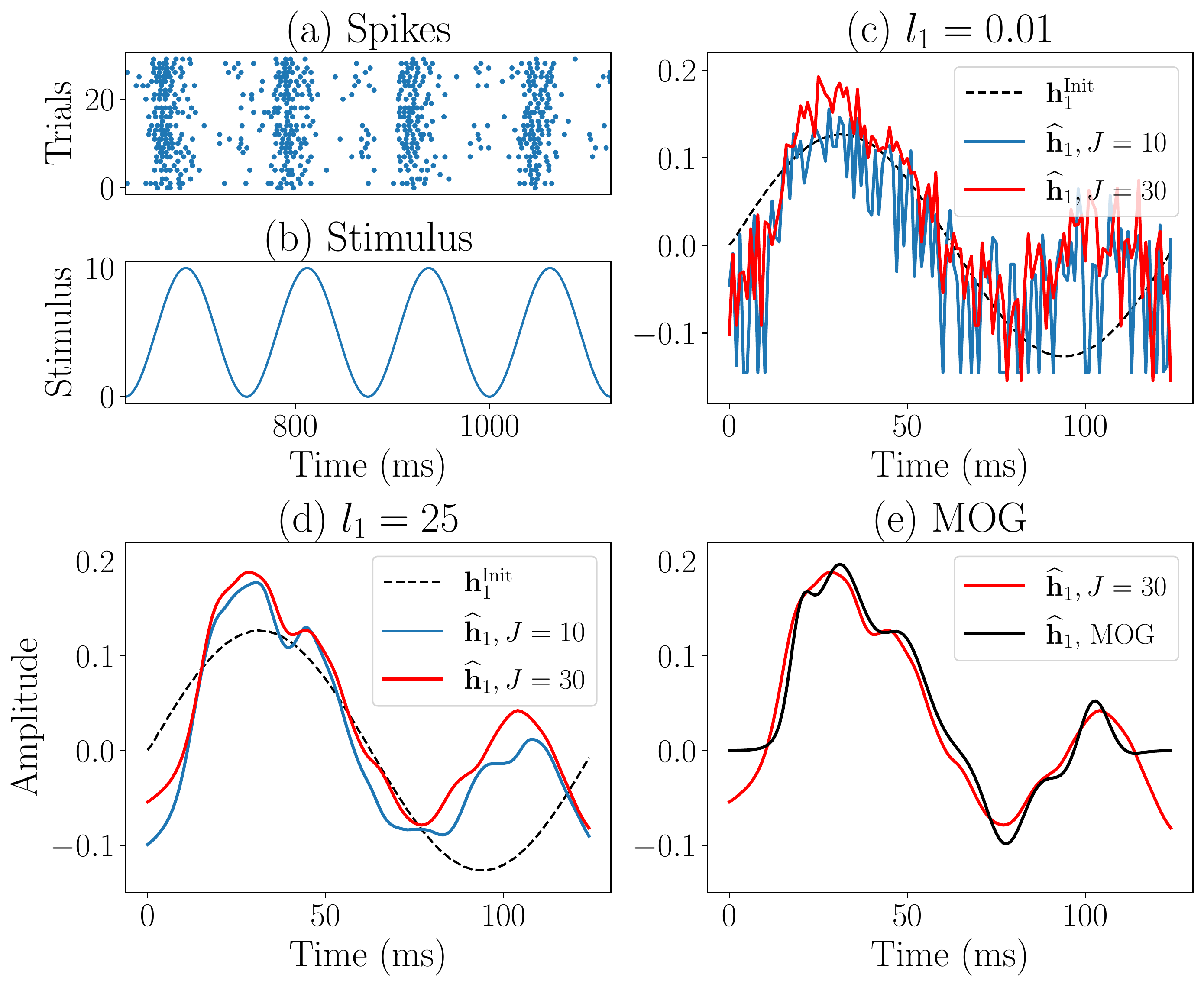}
	\caption{Real data for Neuron 1. (a-b) Raster plot of the spikes and periodic stimulus. (c-d) $\widehat{\h}_1$ with $J\!=\!30$ (red) and $J\!=\!10$ (blue) for various $\l_1$. (e) The parametric baseline, $\widehat{\h}_1^{\text{MOG}}$.}
	\label{fig:real}
\end{figure}

\noindent \textbf{Dataset} We used neural spiking data collected from the barrel cortex of mice~\cite{Temereanca2008}. The experiments consist of multiple trials, with each trial $N=3{,}000$ ms and $\y^j\in\{0,1\}^N$. During each trial, a stimulus (Fig.~\ref{fig:real} (b)) is used to deflect the whisker of a mouse every 125 ms. We set $K=125$ accordingly. Because of the presence of a single stimulus, we assumed $C\!=\!1$ as in~\cite{Tolooshams20}. For $\h_{1}^{\operatorname{Init}}$, we used the first-order difference of the stimulus (dotted black). We used the logit function as the canonical link and set $\phi=1$.  We assumed a constant baseline $\mathbf{a}^j=\mathbf{a},\forall j$ and estimate it from all $J$ segments. We also assumed $\x_1^j=\x_1,\forall j$. 

We used $J\!=\!30$ trials for training and $J_{\text{test}}\!=\!10$ trials for testing for each neuron. We performed 3-fold cross-validation on the training data to find $l_1^{\operatorname{CV}}$ that yields the highest predictive log-likelihood (pll). We used the entire training data to estimate $\widehat{\h}_1$ and $\widehat{\x}_1$. We used pll and $R^2$~\cite{Zhao2017} as performance metrics.

\begin{table}[!t]
	\caption{Metrics (the higher the better) for two neurons with $J=30$. MOG represents the mixture of Gaussians.}
	\label{table:real}
	\centering
	\begin{tabular}{|c|c|c|c|c|c|c|c|c|}
		\hline
		&&\multicolumn{3}{c}{Train}& \multicolumn{4}{|c|}{Test}\\
		\hline
		ID & \diagbox{}{$l_c$}& 0.01 &25&200&0.01&25&200&MOG\\
		\hline
		1&pll&0.57&\textbf{0.60}&0.57&0.61&\textbf{0.65}&0.5&0.62\\
		1&$R^2$&0.28&\textbf{0.30}&0.25&0.27&\textbf{0.30}&0.25&0.29\\
		\hline
		2&pll&0.59&\textbf{0.63}&0.62&0.64&\textbf{0.70}&0.67&0.69\\
		2&$R^2$&0.22&\textbf{0.24}&0.23&0.18&\textbf{0.23}&0.21&0.23\\
		\hline
	\end{tabular}
\end{table}

\noindent \textbf{Results} Table~\ref{table:real} shows the metrics for two neurons. Figs.~\ref{fig:real} (c-d) shows $\widehat{\h}_1$ corresponding to varying $l_1$ for Neuron 1 with $J\!=\!30$ (red). Both the highest pll and $R^2$ for the cross-validation is achieved for $l_1^{\operatorname{CV}}=25$. For the test data, $l_1^{\operatorname{CV}}$ also performs the best. We observe the two peaks in $\widehat{\h}_1$ (red), around 30 and 100 ms, validated by the repeated pattern of the strong bursts of spikes followed by the weak burst. For $l_1\!=\!0.01$, although the two peaks can be identified, $\widehat{\h}_1$ lacks smoothness, as a result of overfitting to the \textit{integer-valued} observations without the smoothness constraint. For $l_1\!=\!200$ with strong regularization, $\widehat{\h}_1$ is overly smoothed and produces lower metrics.

Comparison between $J\!=\!10$ (blue) and $J\!=\!30$ (red) shows the benefits of the regularization for limited data. Without regularization (Fig.~\ref{fig:real} (c)), $\widehat{\h}_1$ for $J\!=\!10$ is noisier than that for $J\!=\!30$ due to the scarcity of data, in addition to the nonlinear link. For $l_c^{\operatorname{CV}}$, $\widehat{\h}_1$ for both cases are similar, showing that the regularized dictionary is robust for limited data.  

Finally, we compared $\widehat{\h}_1$ with $\widehat{\h}^{\text{MOG}}_1$. We chose $D=6$ that minimizes the Akaike Information Criterion~\cite{akaike81}. Fig.~\ref{fig:real} (e) shows that $\widehat{\h}_1^{\text{MOG}}$ is indeed very similar to $\widehat{\h}_{1}$ with $l_1^{\operatorname{CV}}$. However, Table~\ref{table:real} shows that the nonparametric and regularized approaches outperform the parametric alternative, indicating the flexibility of the nonparametric approach. 

\section{Conclusion}\label{section:conclusion}
We proposed a framework for learning convolutional dictionaries using data from the natural exponential family by regularizing the classical objective with a Gaussian process prior. We show that the smoothness constraint leads to a dictionary with better performance. GPCDL is a powerful framework that combines 1) the smoothness previously achieved by parametric functions, which is vulnerable to model misspecification issues, or penalty functions, which are nontrivial to optimize, and 2) the flexibility of the nonparametric dictionary.

\bibliographystyle{IEEEtran}
\bibliography{arxiv}

\clearpage

\section*{Appendix A. Gradient \& Hessian of the loss function}
We discuss how to obtain the gradient and Hessian of the negative log-posterior $\loss$ in Eq.~(2), with respect to $\h_c$. For notational simplicity, we drop dependence on $j$ and $t$. The gradients of the first and the last term, $-\nabla_{\h_c}f(\boldmu)^{\text{T}}\y$ and $\nabla_{\h_c}\h_c^{\text{T}}\Sigma_c^{-1}\h_c$ respectively, are given as follows
\begin{equation*}
\begin{split}
-\nabla_{\h_c}f(\boldmu)^{\text{T}}\y&=-(\X_c)^{\text{T}}\y\\
\nabla_{\h_c}\h_c^{\text{T}}\Sigma_c^{-1}\h_c& = 2\Sigma_c^{-1}\h_c.
\end{split}
\end{equation*}
For the gradient of the second term $\nabla_{\h_c} \mathbf{1}_N^{\text{T}}b(f(\boldmu))$, denoting $\boldeta=f(\boldmu)$ for simplicity, we have the following
\begin{equation*}
\begin{split}
\nabla_{\h_c} \mathbf{1}_N^{\text{T}}b(\boldeta) &= \frac{\partial b(\boldeta)}{\partial \h_c}\mathbf{1}_N\\
&=\frac{\partial \boldeta}{\partial \h_c}\frac{\partial b(\boldeta)}{\partial \boldeta}\mathbf{1}_N\\
&=\X_c^\text{T}\operatorname{diag}(\boldmu)\mathbf{1}_N=\X_c^{\text{T}}\boldmu.\\
\end{split}
\end{equation*}
We use the well-known relationship for the natural exponential family~[16], which states that $d b(\boldeta_i)/d \boldeta_i=\mathbb{E}[\y_i]=\boldmu_i$ with the subscript $i$ referring to $i^{\text{th}}$ element of the corresponding vector. We get $\nabla_{\h_c} \loss$ in Eq.~(3) by collecting these terms.
For the Hessian, we compute $\partial (\X_c^{\text{T}}\boldmu)/\partial \h_c$ as follows
\begin{equation*}
\begin{split}
\frac{\partial\X_c^{\text{T}}\boldmu}{\partial \h_c}&=\frac{\partial \boldmu}{\partial \h_c}\X_c=\frac{\partial f^{-1}(\boldeta)}{\partial \h_c}\X_c\\
&=\frac{\partial \boldeta}{\partial \h_c}\frac{\partial f^{-1}(\boldeta)}{\partial \boldeta}\X_c\\
&=\X_c^{\text{T}}\operatorname{diag}((f'(\boldmu_i))^{-1})\X_c.\\
\end{split}
\end{equation*}

\section*{Appendix B. Maximum likelihood parameter estimation}
Cross-validation for parameter estimation, while easy to evaluate on objective functions of choice without the need for optimization, scales poorly as a function of the number of templates $C$. As an alternative, we can use \textit{approximate} maximum marginal likelihood to estimate the hyperparameters $\theta=\{l_c, \sigma_c^2\}_{c=1}^C$ of the GP kernels. The lack of conjugacy between the likelihood and the prior makes the need for approximation, the details of which are provided below,  necessary. It is exact only when the likelihood is Gaussian.

At step $p$ of the algorithm, we
\begin{enumerate}
	\item perform convolutional sparse coding (CSC) with $\{\h^{(p-1)}_c\}_{c=1}^C$ to obtain sparse codes $\{\X_c^{j,(p)}\}_{c=1}^C$,
	\item perform convolutional dictionary update to obtain $\{\h_c^{(p)}\}_{c=1}^C$, using the GP kernel parameters $\theta^{(p-1)}$,
	\item obtain the marginal likelihood using Laplace approximation around $\{\h_c^{(p)}\}_{c=1}^C$,
	\item compute the gradient with respect to $\theta^{(p-1)}$, and take a gradient ascent step to obtain $\theta^{(p)}$.
\end{enumerate}

Steps $1-3$ produce the approximate marginal log-likelihood $\log p(\{\y^j\}_{j=1}^J \mid \theta^{(p-1)})$ and step 4 performs the gradient ascent step. The steps are repeated until convergence. 

We now expand on step 3, which largely follows the steps in [13], with specific modifications to fit the GPCDL generative model. Denoting $\widetilde{\h} =[(\h_1)^{\text{T}}, \ldots, (\h_C)^{\text{T}}]^{\text{T}}\in\mathbb{R}^{CK}$, a concatenation of all templates, we can use the Laplace approximation on the \textit{unnormalized} posterior $\exp(\Psi(\widetilde{\h}))$ to obtain the marginal likelihood
\begin{equation*}\label{eq:laplace}
\begin{split}
&p(\{\y^j\}_{j=1}^J \mid \theta^{(p-1)})\\
&= \int \underbrace{p(\{\y^j\}_{j=1}^J\mid \{\h_c \}_{c=1}^C)p(\{\h_c \}_{c=1}^C\mid \theta^{(p-1)})}_{\exp(\Psi(\widetilde{\h}))}\prod_{c=1}^Cd\h_c\\
&=\int \exp(\Psi(\widetilde{\h}))d\widetilde{\h}\\
&\simeq\exp(\Psi(\widetilde{\h}^{(p)}))\\
&\quad\times\int\exp(\frac{1}{2}(\widetilde{\h}-\widetilde{\h}^{(p)})^{\text{T}}\nabla^2_{\widetilde{\h}}\Psi(\widetilde{\h}^{(p)})(\widetilde{\h}-\widetilde{\h}^{(p)}))d\widetilde{\h},\\
\end{split}
\end{equation*}
where we perform Laplace approximation on $\Psi(\widetilde{\h})$ by performing Taylor expansion of $\Psi(\widetilde{\h})$ around $\widetilde{\h}^{(p)}$ 
\begin{equation*}
\Psi(\widetilde{\h})\simeq \Psi(\widetilde{\h}^{(p)}) + \frac{1}{2}(\widetilde{\h}-\widetilde{\h}^{(p)})^{\text{T}}\nabla^2_{\widetilde{\h}}\Psi(\widetilde{\h}^{(p)})(\widetilde{\h}-\widetilde{\h}^{(p)}).
\end{equation*}
Note that $\nabla^2_{\widetilde{\h}}\Psi(\widetilde{\h})$ is the same as $-\nabla^2_{\widetilde{\h}}\loss$ of GPCDL. The integral is analytically tractable, which finally yields the approximate marginal log-likelihood
\begin{equation*}
\begin{split}
&\log p(\{\y^j\}_j \mid \theta^{(p-1)})\simeq -\frac{1}{2}\log(\det( \mathbf{I}+\mathbf{B}^{\frac{1}{2}}\Sigma\mathbf{B}^{\frac{1}{2}}))\\
&\quad\quad-\frac{1}{2}(\widetilde{\h}^{(p)})^{\text{T}}\Sigma^{-1}\widetilde{\h}^{(p)} + \sum_{j=1}^J\log p(\y^j\mid \widetilde{\h}^{(p)},\theta^{(p-1)}),
\end{split}
\end{equation*}
where $\Sigma$ is a block diagonal of covariance matrices parametrized by $\theta^{(p-1)}$ and
\begin{equation*}
\mathbf{B}=\phi^{-1}\sum_{j=1}^J(\X^{j,(p)})^{\text{T}}\operatorname{diag}((f'(\boldmu^{j,(p)}))^{-1})\X^{j,(p)},
\end{equation*}
with $\X^{(p)}=[\X_1^{j, (p)},\ldots, \X_C^{j, (p)}]\in\mathbb{R}^{N\times C(N-K+1)}$.

In Step 4, we take the gradient of the approximate marginal log-likelihood $\log p(\{\y^j\}_{j=1}^J \mid \theta^{(p-1)})$ with respect to $\theta^{(p-1)}$ to obtain $\theta^{(p)}$. For more details on the computation of the gradient, we refer the readers to Chapter 5.5.1 of [13].

\end{document}